\documentclass[letterpaper, 10 pt, conference]{ieeeconf}

\IEEEoverridecommandlockouts                              %

\title{\LARGE \bf
Flying Adversarial Patches: Manipulating the Behavior of Deep Learning-based Autonomous Multirotors
}

\author{Pia Hanfeld$^{1,2}$, Marina M.-C. Höhne$^{3,4}$, Michael Bussmann$^{1}$, and Wolfgang Hönig$^{2}$%
\thanks{$^{1}$ CASUS, Helmholtz-Zentrum Dresden-Rossendorf, $^{2}$ Technical University Berlin, $^{3}$ Leibniz Institute for Agricultural Engineering and Bioeconomy, $^{4}$ University of Potsdam}
\thanks{Corresponding author: p.hanfeld@hzdr.de}
\thanks{Code:\url{https://github.com/IMRCLab/flying_adversarial_patch/tree/icra_mrlearn23}}
\thanks{The research was partially funded by the Deutsche Forschungsgemeinschaft (DFG, German Research Foundation) - 448549715. Furthermore, it was partially funded by the Center for Advanced Systems Understanding (CASUS), financed by Germany’s Federal Ministry of Education and Research (BMBF), and by the Saxon state government out of the State budget approved by the Saxon State Parliament.
}%
}

\usepackage[bookmarks=true]{hyperref}
\usepackage{amssymb}
\usepackage{amsmath}
\usepackage{multirow}
\usepackage[detect-weight=true, detect-family=true,detect-inline-weight=math]{siunitx}

\usepackage[detect-weight=true, detect-family=true,detect-inline-weight=math]{siunitx}

\usepackage[ruled,vlined,linesnumbered]{algorithm2e}
\SetInd{0.25em}{0.5em}
\SetAlFnt{\footnotesize\sf}
\SetKwRepeat{Do}{do}{while}
\SetKw{KwStep}{step}
\SetKwInput{KwData}{Input}
\SetKwInput{KwResult}{Result}
\SetKwBlock{RepeatForever}{repeat}{}
\SetKw{Continue}{continue}
\SetKwComment{Comment}{$\triangleright$\ }{}
\SetCommentSty{itshape}

\usepackage[capitalise]{cleveref} %
\usepackage{flushend}

\usepackage[natbib=true,
    style=ieee,
    citestyle=numeric-comp,
    backend=bibtex,
    maxbibnames=10,
    maxcitenames=1,
    mincitenames=1,
    doi=false,
    url=false,
    giveninits=true]{biblatex}
\renewcommand*{\bibfont}{\footnotesize}

\usepackage{tikz}
\usetikzlibrary{fit,calc}
\newcommand{\vp}{\mathbf{p}}    %
\newcommand{\vtheta}{\boldsymbol{\theta}}
\newcommand{\vt}{\boldsymbol{t}}

\newcommand{\mC}{\mathbf{C}}    %
\newcommand{\mP}{\mathbf{P}}    %
\newcommand{\mT}{\mathbf{T}}    %

\newcommand{\sC}{\mathcal{C}}   %
\newcommand{\sS}{\mathcal{S}}   %
\newcommand{\sE}{\mathcal{E}}   %
\newcommand{\sT}{\mathcal{T}}   %

\newcommand{\vf}{\mathbf{f}}    %
\DeclareMathOperator*{\place}{place}
\DeclareMathOperator*{\ADAM}{ADAM}
\DeclareMathOperator*{\random}{random}

\DeclareMathOperator*{\argmin}{argmin}

\begin{document}

\maketitle
\thispagestyle{empty}
\pagestyle{empty}

\begin{abstract}

Autonomous flying robots, e.g. multirotors, often rely on a neural network that makes predictions based on a camera image. These deep learning (DL) models can compute surprising results if applied to input images outside the training domain. Adversarial attacks exploit this fault, for example, by computing small images, so-called adversarial patches, that can be placed in the environment to manipulate the neural network's prediction. We introduce flying adversarial patches, where an image is mounted on another flying robot and therefore can be placed anywhere in the field of view of a victim multirotor. For an effective attack, we compare three methods that simultaneously optimize the adversarial patch and its position in the input image. We perform an empirical validation on a publicly available DL model and dataset for autonomous multirotors. Ultimately, our attacking multirotor would be able to gain full control over the motions of the victim multirotor.
\end{abstract}

\section{Introduction}

Deep learning (DL) models are susceptible to adversarial attacks~\cite{Szegedy2014, Goodfellow14}. Applying tiny adversarial perturbations to each input can lead to misclassifications by the neural Network (NN), negatively affecting the NN's accuracy. Still, DL models are widely applied in critical infrastructure, e.g., for the perception and control of Unmanned Aerial Vehicles (UAVs). Manipulating the NN's input data to include the calculated adversarial perturbations usually requires direct access to the data, rendering classic adversarial attacks as infeasible for a potential real-world attacker. 
Adversarial patches are special adversarial attacks on image classification models that can be printed and placed anywhere in a real-world environment~\cite{Brown2017, Sharma2022}.

With the help of DL models implemented in autonomous multirotors, they can track and follow objects or human subjects in their environment. Tracking subjects is usually performed with estimations of the position and orientation angle, i.e. the \textit{pose}, of the subject. For pose estimation, the UAVs are equipped with a camera, and the camera images are then forwarded to a DL model predicting the pose relative to, e.g., the UAV's frame of reference. These poses can be utilized to generate new waypoints for the UAV, enabling the UAV to follow the subject. 

\begin{figure}
    \centering
    \includegraphics[width=0.5\textwidth]{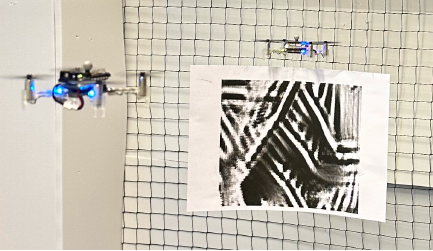}
    \caption{The attack scenario of this work: a multirotor carries an adversarial patch. The adversarial patch is able to prevent the NN from predicting the relative pose of a human subject in the input image, regardless of the human's position in the image. Instead, the NN predicts a pose previously chosen by an attacker.}
    \label{fig:overview}
\end{figure}

A second mutlirotor--referred to as the attacker UAV--will execute the attack of this work. The computed adversarial patch can be printed and affixed to the attacker UAV (see \cref{fig:overview}). 
If the predictions of the DL model are manipulated successfully by the placed adversarial patches, the attacker can gain full control over the UAV--referred to as the victim UAV.

Since the attacker UAV can only carry a limited number of patches, one patch needs to be applicable to multiple positions in the image.

In this paper, we focus on describing the optimization of an adversarial patch and its position in the input image.

The contributions of this paper are as follows:
\begin{enumerate}
    \item We present novel algorithms to optimize an adversarial patch and its position as a targeted adversarial attack.
    \item We demonstrate the functionality of our method on a publicly available NN and a dataset for UAVs.
\end{enumerate}

\section{Related Work}
Adversarial attacks can be grouped into two categories: white-box attacks, when using the known network parameters, and black-box attacks if the network's parameters are unknown.
Adversarial perturbations can be calculated to be applicable on a single input image only or on most images within a given image dataset, referred to as universal adversarial perturbation (UAP)~\cite{Moosavi-Dezfooli2017}.
Typically, classic (universal) adversarial perturbations are limited to be invisible to the human eye to achieve a stealthy attack. Adversarial patches are usually not hidden and appear like stickers or murals.~\cite{Eykholt2018} 

Adversarial patches are special universal adversarial perturbations that can be printed and placed anywhere in the environment to be visible in the input images while maintaining a severe negative impact on the NN's performance. Adversarial patches are, therefore, a versatile tool to manipulate the prediction of a DL model without the need to manipulate the input image on a pixel level.

So far, many adversarial attacks focus on autonomous driving~\cite{Eykholt2018, Tu2020, Cao2022, Xu2020}. In the following, we focus on adversarial attacks for UAVs.

\citet{Raja2021} introduce a white- and black-box attack algorithm to an image classification NN deployed onboard of a UAV for bridge inspection. The NN's task is to identify risk-prone regions in the images. The output of the NN, therefore, does not influence the control of the UAV. The authors restrict the generated adversarial patches to be placed on physical objects only. They search within the whole image space for the optimal positions for the patches. The colors of the generated patches are additionally limited to stay in the set of printable colors.

\citet{Tian2021} introduce two white-box, single-image adversarial attacks on the publicly available DroNet~\cite{loquercio2018dronet}. Both methods can impact the steering angle and collision probability predicted by the NN. Their attacks force the network's predictions to be as far away as possible from the original predictions--they perform a so-called untargeted attack. In this case, the attacker has no control over the outcome of the attack and might even improve the performance of the NN. 

\citet{Nemcovsky2022} perform an adversarial patch attack on the Visual Odometry system used for navigation in simulation and real-world experiments. The patches were placed at fixed positions in both environments utilizing Aruco markers. In both examples, they were able to show that the attacks had a severe negative impact on the predicted traveled distance. Other than the previously described works, they take the influence on a sequence of images and different view angles into account.

In this work, we go beyond current approaches and introduce a targeted, white-box adversarial patch attack, that enables a potential real-world attacker to gain full control over a victim UAV. Instead of fixing the patches to certain positions, we additionally optimize a single patch to be applicable to multiple positions in the camera images. Furthermore, our method includes the optimization of these positions, to provide the attacker with an effective attack policy.

\section{Background}

In the following, we introduce the NN we attack, PULP-Frontnet, a DL model that tracks a human and has been deployed on small multirotors. Moreover, we provide an introduction to adversarial patch placement.

\subsection{PULP-Frontnet}

PULP-Frontnet~\cite{Palossi2022} is a publicly available DL model implemented in PyTorch and developed for a nano multirotor--the Crazyflie by Bitcraze\footnote{\url{https://www.bitcraze.io/products/crazyflie-2-1/}}. The NN performs a human pose estimation task on the input images, predicting a human subject's 3D position and yaw angle in the UAV's frame. This prediction is later used in the UAV's controller to generate new setpoints which are subsequently used to follow the human subject. Therefore, the prediction of PULP-Frontnet directly influences the UAV's behavior.
For the generation of the adversarial patches of this work, we utilized the testset provided by \cite{Palossi2022}. The testset, referred to as $\sC$, consists of 4028 images of size $\mC_i\in\mathbb{R}^{160 \times 96}$ cropped at the same vertical position and without further augmentations, unlike the training dataset. This testset provides images that are very close to a setting for a real-world attack.
The forward pass of the NN $\vf_{\vtheta}(\cdot)$ can be defined as
\begin{align}
    \hat{\vp}^h &= \vf_{\vtheta}(\mC)\,,
\end{align}
where $\vtheta$ are the parameters of the NN, and the output is the estimated pose of the human subject $\hat{\vp}^h = (\hat{x}^h, \hat{y}^h, \hat{z}^h, \hat{\phi}^h)^T$ in the UAV's coordinate frame, where  $\hat{x}^h$ determines the distance in meters (depth), $\hat{y}^h$ the horizontal, $\hat{z}^h$ the vertical distance, and $\hat{\phi}^h$ the orientation angle of the subject to the UAV.
The controller of PULP-Frontnet uses $\hat{\vp}^h$ to compute a setpoint that keeps the human in the camera view. Thus, the UAV will follow the human subject depending on the predicted $\hat{\vp}^h$.
Therefore, manipulating the neural network output will directly influence the motions of the UAV.
\begin{figure*}
    \centering
    \includegraphics[width=0.95\textwidth]{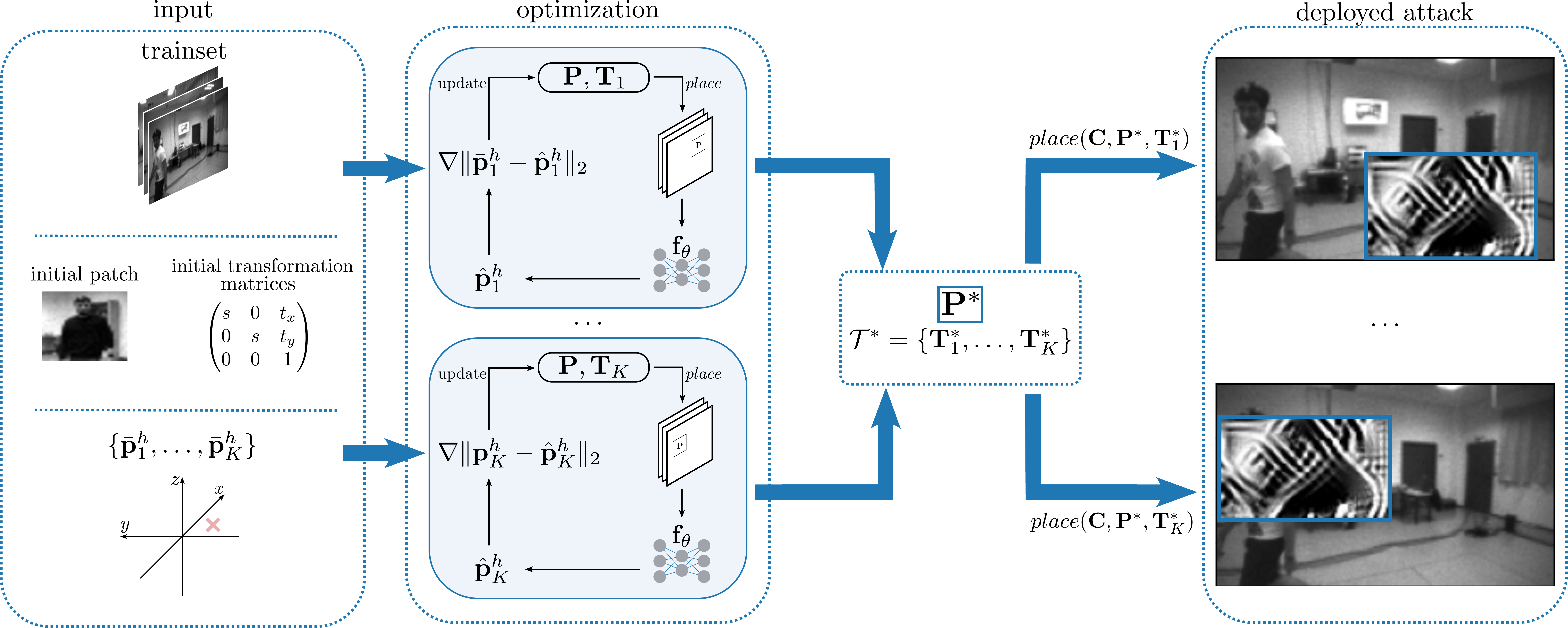}
    \caption{We compute i) an optimal adversarial patch $\mP^*$ and ii) a set of optimal transformation matrices $\sT^*$, such that an attacker can place $\mP^*$ in any image from the dataset $\sC$ and the error for the $k^{\text{th}}$ target position $\bar{\vp}^h_k$ and the predicted pose $\hat{\vp}^h_k$ under the attack is minimal. Left: The inputs are the trainset $\sS$, the initial guesses for $\mP$ and $\sT$, and the set of target positions $\{\bar{\vp}^h_1, \ldots, \bar{\vp}^h_K\}$. Middle: $\mP$ and $\sT$ are optimized such that $\|\bar{\vp}^h_k- \hat{\vp}^h_k\\|_2$ becomes minimal for all $K$ target positions. The results are the optimal patch $\mP^*$ and optimal set of transformation matrices $\sT^*$. Right: $\mP^*$ (marked with a blue frame) can be placed in all images given the set of transformations $\sT^*$. The manipulated images are utilized as input to the NN, and the attacker gains full control over the NN's predictions.}
    \label{fig:scetch_approach}
\end{figure*}
\subsection{Differentiable Adversarial Patch Placement}\label{sec:fixed_approach}
\citet{Thys2019} introduce an adversarial patch attack that prevents the detection of a human subject in camera images. They perform their attack on YOLOv2~\cite{redmon2017yolo9000}, a common NN for object detection, trained on the Microsoft COCO dataset~\cite{lin2014microsoft}. It predicts bounding boxes and classes for each object in an input image. Here, the main goal of the attack is to minimize the classification and a so-called objectness score for the subject. Minimizing the class score leads to the model misclassifying the subject as another class. The objectness score indicates the likelihood of an object in the detected region.
Similarly to~\cite{Raja2021}, they restrict the patches to be placed on humans in the images. To improve robustness towards different view angles, lighting conditions, rotation, and scaling, they place and transform their patches with a method inspired by Spatial Transformers Networks~\cite{jaderberg2015spatial}.
The patches are placed and transformed with an affine transformation given a randomly generated transformation matrix. Additionally, they add random noise to counter the influence of noise added by the printer, the camera, and the lighting conditions. %

The affine transformation matrix is defined as \begin{align}
    \mT &= \begin{pmatrix}
        s \cos\alpha & -s\sin\alpha & t_x \\
        s\sin\alpha & s\cos\alpha & t_y \\
        0 & 0 & 1
    \end{pmatrix}\,,
\end{align}
where $s$ is a scale factor, $\alpha$ is the rotation angle, and $ \vt = (t_x,  t_y)^T$ is the translation vector.

Let $\mP\in\mathbb{R}^{w \times h}$ be a grayscale adversarial patch with width $w$ and height $h$. The patch can be placed in image $\mC$
\begin{align}
    \mC' &= \place(\mC, \mP, \mT)\,,
\end{align}
where $\place$ is the differentiable placement function and $\mC'$ is the resulting modified image. Note that the whole transformation process is differentiable.

In this work, we propose to additionally optimize the parameters of the transformation matrix, e.g., the scale, rotation, and translation vector instead of randomly choosing them during the training process. Optimizing those parameters will provide an attack policy, leading to a more effective and realistic attack scenario.

\section{Approach}\label{sec:approach}

The main goal of our method is to find i) an adversarial patch and ii)  a policy for the attacker. The policy maps the current and the desired positions of the victim UAV to the best transformation matrix from a set of optimized matrices. Placing the adversarial patch with the chosen transformation matrix can be achieved by moving the attacker relative to the victim.
The placed patch changes the NN output to the desired pose, such that the attacker gains full control over the victim UAV.

We set the rotation angle of all transformations $\mT$ to $\alpha = 0$, because rotation angles $\alpha \neq 0$ would require specifying a desired roll or pitch angle of the attacker UAV, which is not a controllable degree of freedom for a standard multirotor.

We split the dataset into two disjoint parts $\sC=\sS \cup \sE$ for training ($\sS$) and testing ($\sE$), respectively.

\subsection{Problem Definition}

Given are the target prediction ($\bar{\vp}^h$) and the current position of the human subject ($\vp^h$).
Our goal is to obtain i) a patch $\mP$, and ii) a policy for the attacker that computes the desired pose of the attacking UAV, given the current and desired positions of the victim: $\bar{\vp}^a = \pi^a(\vp^h, \bar{\vp}^h)$.
The objective is to minimize the tracking error of the victim between its current pose $\hat{\vp}^v$ and desired pose $\bar{\vp}^v$, i.e. $\int_t \| \bar{\vp}^v(t) - \hat{\vp}^v(t)\| dt$.

In this paper we manually specify $\pi^a$: the attacker UAV moves such that the patch it carries will be visible in the victim camera frame with a specified transformation $\mT$ (translation and scaling). Then the above can be simplified as follows.
Given are the $K$ target positions $\{\bar{\vp}^h_1, \ldots, \bar{\vp}^h_K\}$. Our goal is to compute i) a patch $\mP$, and ii) a set of transformations $\sT=\{\mT_1, \ldots, \mT_K\}$ (one for each target position). %
The objective is to minimize
\begin{align}
    \label{eq:loss}
    L = & \frac{1}{|\sS|}\sum_{\mC\in\sS} \sum_{\mT_k\in\sT} \|\bar{\vp}^h_k- \vf_\theta(\place(\mC, \mP, \mT_k))\|_2\;.
\end{align}
There exists an optimal adversarial patch $\mP^*$ and an optimal set of transformation matrices $\sT^*$, such that $\mP^*, \sT^* = \argmin_{\mP, \sT} L\,.$

For each target position of the subject $\bar{\vp}^h_k$, the optimal adversarial patch $\mP^*$ can be placed at the position given $\mT_k^*$ to receive $\mC'_k$, such that the error between the prediction $\vf_{\theta}(\mC'_k)$ and the target position $\bar{\vp}^h_k$ is minimized. The whole procedure is visualized in \cref{fig:scetch_approach}.

\subsection{Joint Optimization}\label{sec:joint}

One approach is the simultaneous optimization of the adversarial patch and the transformation matrix. %
Hence, $\mP$ and $\sT$ are updated after each iteration using the gradient of \cref{eq:loss} with respect to $\mP$ and $\sT$: 
\begin{align}
    \mP, \sT \gets \ADAM(\nabla_{\mP, \sT} L).
\end{align}

Note that the parameters $\vtheta$ of the NN $\vf$ are fixed and not changed during the optimization.
To increase the numerical robustness of $\mP$ and $\sT$, we add small random perturbations to each image after the patch placement and to each $\mT_k$, using a normal distribution.

\subsection{Split Optimization}
The joint optimization might lead to a local optimum instead to the global solution for the optimal position of the adversarial patch due to the local nature of the gradient-based optimization. Therefore, we propose to split the training algorithm into two parts: Inspired by the expectation-maximization (EM) algorithm~\cite{dempster1977maximum}, we first only optimize the adversarial patch $\mP$ for a fixed set of transformations $\sT$. For a user-defined number of iterations $N$, the patch is optimized for the $K$ desired positions $\{\bar{\vp}^h_1, \ldots, \bar{\vp}^h_K\}$, while the transformation matrices $\sT$ are fixed, see \Cref{alg:split}, \cref{alg:split:P_only_start} - \cref{alg:split:P_only_end}. 

As a second step, we perform \textit{random restarts} (from \cref{alg:split:T_begin} in \Cref{alg:split}). Not only are the current parameters of the matrix $\mT_k$ optimized but, for $R-1$ restarts, chosen randomly and trained for an iteration over $\sS$ (\cref{alg:split:restarts} - \cref{alg:split:update_T}). We then choose the transformation matrix $\mT$ that produces the minimal loss over all restarts $R$ (\cref{alg:split:update_mTk}), and finally update the current set $\sT$ with the optimized transformation matrices (\cref{alg:split:update_sT}). These random restarts aim to overcome local minima while searching for effective transformation matrices over the whole training process.

\begin{algorithm}[t]
    \caption{Split Optimization}
    \label{alg:split}
    \DontPrintSemicolon
     \KwData{$N$, $R$,  $\sS$, $\mP$, $\sT$, $\vf_\theta$, $\{\bar{\vp}^h_1, \ldots, \bar{\vp}^h_K\}$}
     \KwResult{$\mP^*$, $\sT^*$}

\ForEach{$n \in \{1, \ldots N\}$}{\label{alg:split:P_only_start}
    \ForEach{$k\in\{1, \ldots, K\}$}{
     $L \gets \frac{1}{|\sS|}\sum_{\mC\in\sS} \sum_{\mT_k\in\sT} \|\bar{\vp}^h_k-\vf_{\vtheta}(\place(\mC, \mP, \mT_k)\|_2$ %
     }

    $\mP \gets \ADAM(\nabla_{\mP}L)$   \Comment*{update $\mP$}
    \label{alg:split:P_only_end}
     
     \ForEach{$k\in\{1, \ldots, K\}$}{
     \label{alg:split:T_begin}
         \ForEach{$r \in \{1, \ldots, R\}$}{
         \label{alg:split:restarts}
            \eIf{$r = 1$}{
                 $\mT \gets \mT_k$ \Comment*{optimize the current $\mT_k$}}{
                $\mT \gets \random\mT()$\Comment*{or a randomly initialized $\mT$}
                }
            \label{alg:split:restarts_end}
        $L_r \gets \frac{1}{|\sS|}\sum_{\mC\in\sS}\|\bar{\vp}^h_k-\vf_{\vtheta}(\place(\mC, \mP, \mT)\|_2$\label{alg:split:loss_restart}\\
        $\mT_r \gets \ADAM(\nabla_{\mT}L_r)$\label{alg:split:update_T} \Comment*{update $\mT$}
      }
      $\mT_k \gets \argmin_{\mT_r} \{L_1, \ldots, L_R\}$\label{alg:split:update_mTk} \Comment*{choose best $\mT$}
    }
    $\sT \gets \{\mT_1, \ldots, \mT_k\}$\label{alg:split:update_sT}
}
\Return $\mP^*$, $\sT^*$
\end{algorithm}

\subsection{Hybrid Optimization}
The hybrid version of our method combines the joint and split optimization. The patch and the transformation matrix are trained jointly for a fixed amount of iterations over $\sS$. The parameters of the transformation matrix $\mT_k$ are then fine-tuned while the patch $\mP$ is fixed, analogous to the optimization described in \Cref{alg:split} starting from \cref{alg:split:T_begin}.

\section{Results}
We now analyze all of our proposed approaches, described in \cref{sec:approach}, and compare them to the method introduced by~\cite{Thys2019}, referred to as \textit{fixed optimization}.
We implement our method in PyTorch, utilizing the Adam optimizer~\cite{Kingma2014} to perform the gradient-based optimization. We compute all optimization approaches in parallel on compute nodes with access to 4 cores of Intel 12-Core Xeon (3,0 GHz) CPUs and an NVIDIA Tesla P100 GPU respectively.
The learning rate for all approaches is set to $0.001$, regardless of which combination of parameters is optimized. The trainset $\sS$ contains $90\%$ of the images from the dataset $\sC$. Batches of size 32 are drawn uniformly at random from $\sS$ in each training step, and we train on all batches for one epoch. 
The training process is repeated ten times with random initial seeds, referred to as \textit{trials}. For each approach, $\mP$ and $\sT$ are trained for $N=100$ iterations on the trainset $\sS$. Before placing $\mP$, we add Gaussian noise with a standard deviation of $0.1$ to all matrices in $\sT$. Additionally, we add Gaussian noise to the manipulated images $\mC'$ with a standard deviation of $10$.
To ensure that the patches are visible in $\mC'$, we restrict the parameters of the transformation matrices in $\sT$. The scale factor $s$ is kept in $[0.3, 0.5]$, the translation vector $\vt$ is kept in $(-1, 1)$.
The number of random restarts for the split and hybrid optimization is $R=20$. We validate the negative impact on the NN $\vf_{\vtheta}$ with the mean of the loss defined in \cref{eq:loss} on the unseen testset $\sE$ over the ten trials. The testset $\sE$ contains ten percent of the images in $\sC$.

\subsection{Comparison of the Different Approaches}
\begin{figure}
    \centering
    \includegraphics[page=22, width=0.23\textwidth, trim= 2 0 2 0, clip]{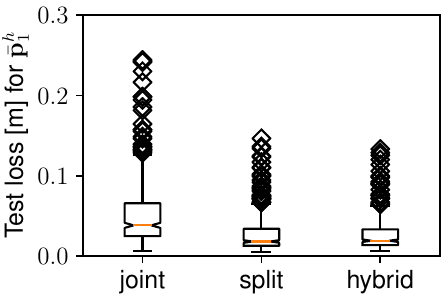}
    \includegraphics[page=23, width=0.23\textwidth, trim= 2 0 2 0, clip]{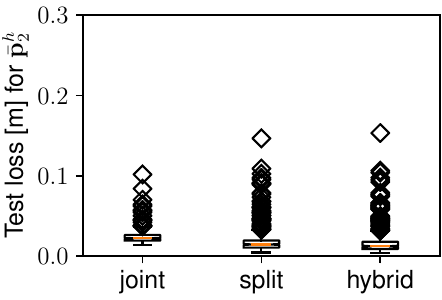}
    \caption{Resulting mean test loss for the proposed joint, split, and hybrid optimization. The joint optimization calculates less effective $\mP$ and $\sT$ for $\bar{\vp}^h_1$ compared to the split and hybrid optimization. The hybrid optimization outperforms the split optimization for both targets slightly.}
    \label{fig:exp1:all}
\end{figure}
\begin{figure}
    \centering
     \includegraphics[page=21, width=0.4\textwidth]{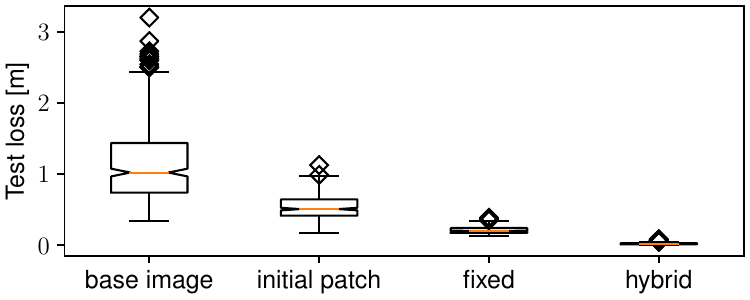}
     \caption{Resulting mean test loss for all $\mC\in\sE$ (i.e. \textit{base image}) per target. For comparison, the \textit{initial patch} $\mP$ (see \cref{fig:scetch_approach}) is placed with the optimal $\sT$ calculated with the hybrid approach. The best results were achieved with the hybrid optimization. Optimizing the adversarial patch $\mP$ with the fixed approach leads to a less effective attack compared to the other optimization approaches.}
     \label{fig:exp1:basevsstartvshybrid}
\end{figure}
We compare the proposed joint, split, and hybrid optimization introduced in \cref{sec:approach} to the fixed optimization. In the fixed optimization case, only the adversarial patch $\mP$ is optimized--the transformation matrix is initialized randomly and fixed for all training steps. We use $K=2$ and the two target poses are $\bar{\vp}^h_1 = (1, -1, 0)^T$ and $\bar{\vp}^h_2 = (1, 1, 0)^T$.

The computed $\mP^*$ and $\sT^*$ utilizing the hybrid optimization on average leads to a $42\%$ lower loss compared to the joint optimization and a $4\%$ lower loss compared to the split optimization on the testset and is, therefore, the best approach for the described setup. The mean test loss over 10 trials for $\bar{\vp}^h_1$ and $\bar{\vp}^h_2$ is displayed in \cref{fig:exp1:all}. The fixed optimization computes a $\mP^*$ that is not as applicable to all images from the testset compared to the other approaches, see \cref{fig:exp1:basevsstartvshybrid}. %

Due to the high number of random restarts $R$ performed during the split and hybrid optimization, the computation time on our hardware setup for 100 iterations over $\sS$ roughly takes 30 times longer than the joint optimization.

\subsection{Ablation Study}
\begin{figure}
    \centering
    \includegraphics[width=0.45\textwidth]{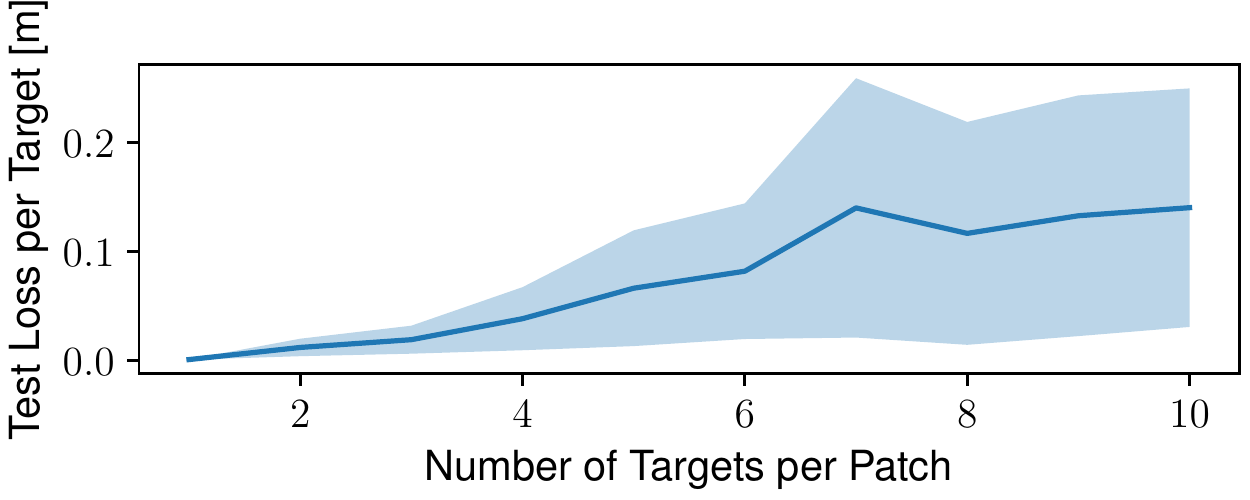}
    \caption{Mean test loss per number of targets over three trials. The shaded area depicts the variance between the trials. The attack is very effective for $1\leq K \leq 4$. The variance becomes larger as soon as target poses $\bar{\vp}^h_k$ are included, which are underrepresented in the original trainset of PULP-Frontnet.}
    \label{fig:exp2}
\end{figure}

We investigate the scalability of our approach with respect to the number of targets $K$, where the results are shown in \cref{fig:exp2}. The first nine targets are set to $\bar{\vp}^h_k = (1, y, z)^T$, where $y, z \in \{-1, 0, 1\}$, representing predictions in the corners and the center of the image $\mC$. The last target is set to $\bar{\vp}^h_10 = (2, 0, 0)$, representing a prediction in the center of the image but further away from the victim UAV.

The variance increases with the number of targets because not all target positions can be equally easily achieved. The loss increases as soon as target poses $\bar{\vp}^h_k$ are included, which are underrepresented in the original trainset of PULP-Frontnet--especially $z > 0.5$ and $z < -0.5$. Additionally, $y < 0$ are more challenging for the adversarial patch attack since the human subjects included in the images of $\sC$ are primarily located at $y > 0$. %

In the previous experiments, we initialized $\mP$ with a human subject included in the original trainset of PULP-Frontnet (see \cref{fig:scetch_approach}). We assumed that a human or a face would have the most influence on the NN, and therefore, the attack is more effective. In \cref{fig:exp1:basevsstartvshybrid}, one can see that the starting patch already negatively influences the NN if placed at the optimal $\sT$. The optimized patches, however, can not be interpreted as human-looking. Therefore, we repeated the training with different starting patches, see \cref{tab:exp3}. The first three patches showed faces from the dataset $\sC$. Although the final, optimal patches $\mP^*$ apparently do not resemble human figures, the imparted prior knowledge starting from a human face improves the negative effect on the NN.
\begin{table}
    \centering
    \caption{Ablation study for different initial patches. The training was repeated for ten trials. Lower is better.}
    \label{tab:exp3}
    \begin{tabular}{l|c|c|c|c|c}
                      & face 1 & face 2 & face 3 & white & random\\\hline
      best approach   &  hybrid & split &  split & joint & hybrid  \\\hline
      mean test loss  &  \textbf{0.024} & 0.052 &  0.040 & 0.099 & 0.053  \\\hline
      mean std        &  \textbf{0.008} & 0.025 &  0.032 & 0.065 & 0.040
    \end{tabular}
\end{table}

\section{Conclusion}
This paper introduces new algorithms to optimize an adversarial patch and its position simultaneously. The adversarial patch allows for full control over the NN's predictions. It can be placed in the attacked image at multiple positions, successfully shifting the predictions of the NN to desired values. %

Our empirical studies show that the described hybrid optimization approach outperforms the joint and split optimization. All introduced approaches exceed the state-of-the-art method, i.e., fixed optimization, for the described attack scenario. Initializing the adversarial patch to resemble a human figure additionally enhances the negative effect of the attack in comparison to a white or random patch.

We observe that the random restarts included in the split and hybrid optimization only find more beneficial $\mT$ during the first few iterations. Since the joint optimization still computes effective adversarial patches and transformation matrices $\sT$, it might be the preferred method given the current use case. The random restarts can be parallelized to increase computation speed. 
The calculated adversarial patch $\mP^*$ can be placed at multiple desired target positions. Depending on how well the predictions of the NN can achieve the desired target positions, the best results will be achieved for $K\leq4$ desired target positions. 

In the future, we will demonstrate the attack in the real world. Therefore, the method needs to be adapted to produce printable adversarial patches. Furthermore, the attack policy can be extended to include multiple patches. Additionally, the hyperparameters for the experiments can be tuned by performing a hyperparameter search. PULP-Frontnet can be deployed on the hardware as a quantized NN. It would be intriguing to attack the quantized version of the NN directly. %
\printbibliography

\end{document}